\documentclass{article}

\usepackage{neurips_2024}

\usepackage[utf8]{inputenc} 
\usepackage[T1]{fontenc}    
\usepackage{hyperref}       
\usepackage{xurl}            
\usepackage{booktabs}       
\usepackage{amsfonts}       
\usepackage{nicefrac}       
\usepackage[none]{hyphenat}
\usepackage{microtype}      
\usepackage{xcolor}         
\usepackage{natbib}         
    \setcitestyle{round}
    \setcitestyle{aysep={}}
\usepackage[symbol]{footmisc}

\title{Toward an evaluation science for generative AI systems}

\author{
  \textbf{Laura Weidinger$^*$} \\
  Google DeepMind \\
  \\
  \textbf{Inioluwa Deborah Raji}\thanks{ = contributed equally to this work}\\
  University of California, Berkeley \\
  \\
   \textbf{Hanna Wallach} \\
  Microsoft Research  \\
    \\
  \textbf{Margaret Mitchell} \\
  Hugging Face \\
    \\
   \textbf{Angelina Wang}\\
  Stanford University \\
    \\
  \textbf{Olawale Salaudeen}\\
  Massachusetts Institute of Technology \\
    \\
   \textbf{Rishi Bommasani}\\
  Stanford University  \\
    \\
   \textbf{Deep Ganguli} \\
  Anthropic  \\
    \\
  \textbf{Sanmi Koyejo}\\
  Stanford University  \\
    \\
   \textbf{William Isaac}\\
  Google DeepMind 
 }

\begin{document}
\maketitle

\begin{abstract}
There is an increasing imperative to anticipate and understand the performance and safety of generative AI systems in real-world deployment contexts. However, the current evaluation ecosystem is insufficient: commonly used static benchmarks face validity challenges, and ad hoc case-by-case approaches rarely scale. In this piece, we advocate for maturing an evaluation \textit{science} for generative AI systems. While generative AI creates unique challenges for system safety engineering and measurement science, the field can draw valuable insights from the development of safety evaluation practices in other fields including transportation, aerospace, and pharmaceutical engineering. In particular, we present three key lessons: evaluation metrics must be applicable to real-world performance, metrics must be iteratively refined, and evaluation institutions and norms must be established. Applying these insights, we outline a concrete path toward a more rigorous approach for evaluating generative AI systems. 
\end{abstract}

\setcounter{footnote}{0}
\section{Introduction}

The widespread deployment of generative AI systems in medicine \citep{boyd_microsoft_2023}, law (e.g., Lexis+AI\footnote{https://www.lexisnexis.com/en-us/products/lexis-plus-ai.page}), education \citep{singer_new_2023}, information technology (e.g., Microsoft's Copilot \footnote{copilot.microsoft.com}, \citet{reid_generative_2024}, and many social settings (e.g. Replika\footnote{replika.com}, character.AI) has led to a collective realization: the performance and safety of generative AI systems in real-world deployment contexts are very often poorly anticipated and understood \citep{roose_.i._2024, mulligan_way_nodate, wiggers_ai_2024}. The tendency of these systems to generate inaccurate statements has already led to the spreading of medical and other misinformation \citep{omiye_large_2023, archer_representation_2025}; incorrect legal references \citep{magesh_ai_2024}; failures as educational support tools \citep{singer_classrooms_2023}; and widespread confusions in search engine use \citep{heaven_why_2022, murphy_kelly_microsofts_2023}. Beyond factual discrepancies, AI-enabled chatbots have also been described as interacting inappropriately with users \citep{roose_conversation_2023}, exposing security vulnerabilities \citep{nicholson_bing_2023}, and fostering unhealthy emotional reliance \citep{roose_can_2024, dzieza_confusing_2024, verma_they_2023}.

The historical focus on benchmarks and leaderboards has been effective at encouraging the AI research community to pursue shared directions; however, as AI products become widely integrated into our everyday lives, it is increasingly clear that static benchmarks are not well suited to improving our understanding of the \textit{real-world} performance and safety of deployed generative AI systems \citep{bowman_what_2021, goldfarb-tarrant_intrinsic_2021, liao_are_2021, raji_bodies_2021, de_vries_towards_2020}. Despite this mismatch, static benchmarks are still commonly used to inform real-world decisions about generative AI systems that stretch far beyond the research landscape -- such as in deployment criteria and marketing materials for new model or system releases \citep{grattafiori_llama_2024, anthropic_introducing_2024, noauthor_gpt-4o_nodate, gemini_team_gemini_2024}, third party critiques \citep{mirzadeh_gsm-symbolic:_2024, zhang_careful_2024}, procurement guidelines \citep{johnson_legacy_2025}, and in public policy discourse \citep{EC_second_2024,NIST_ai_2021}. Although there is an emerging interest in more interactive \citep{chiang_chatbot_2024}, dynamic \citep{kiela_dynabench:_2021}, and behavioral \citep{ribeiro_beyond_2020} approaches to evaluation, many of the existing alternatives to benchmarks, such as red teaming exercises and case-by-case audits, still fall woefully short of enabling systematic assessments and accountability \citep{birhane_ai_2024, friedler_ai_2023}.

For AI evaluation to mature into a proper “science,” it must meet certain criteria. Sciences are marked by having theories about their targets of study, which can be expressed as testable hypotheses. Measurement instruments to test these hypotheses must provide experimental consistency (i.e., reliability, internal validity) and generalizability (i.e., external validity). Finally, sciences are marked by iteration: Over time, measurement approaches and instruments are refined and new insights are uncovered. Collectively, these properties of sciences contrast sharply with the practice of rapidly developing static benchmarks for evaluating generative AI systems, while anticipating that within a few months such benchmarks will become much less useful or obsolete. 

As generative AI exits an era of research and enters a period of widespread use \citep{hu_chatgpt_2023, reid_generative_2024}, the field risks exacerbating an ongoing public crisis of confidence in AI technology \citep{faverio_what_2023} if we do not develop a more mature evaluation science for generative AI systems. From the history of other fields, we can get a sense of why: Collectively, leaderboards, benchmarks, and audits do not amount to the robust and meaningful evaluation ecosystem we need to properly assess the suitability of these products in widespread use. In particular, they cannot give assurances about AI system performance in different domains or for different user groups\footnote{We thank our anonymous reviewer for pointing out that one reason for this is that benchmarks are often not robust to data or domain shifts, i.e. benchmarks test AI system outputs in certain contexts but this may not be predictive of AI system behaviour in other contexts.}. In this piece, we advocate for the maturation of such an evaluation science. By drawing on insights from systems safety engineering and measurement science in other fields, while acknowledging the unique challenges inherent to generative AI, we identify three important properties of any evaluation science that the AI community will need to focus on to meaningfully advance progress: a focus on real-world applicability, iterative measurement refinement, and adequate institutional investment. These properties then enable us to outline a concrete path toward a more rigorous evaluation ecosystem for generative AI systems.

\section{Lessons from Other Fields}

The bridges we stand on, the medicine we take, and the food we eat are all the result of rigorous assessment. In fact, it is because of the rigor of the corresponding  evaluation ecosystems that we can trust that the products and critical infrastructure surrounding us are performant and safe. Generative AI products are no exception to this reality and therefore not unique in their need for robust evaluations. In response to their own crises, more established evaluation regimes emerged in other fields to assure users and regulators of safety and reliability - offering concrete lessons for the AI field \citep{rismani_plane_2023, raji_concrete_2023, raji_fallacy_2022}. We note three key evaluation lessons from these other fields: the targeting of real-world performance, the iterative refinement of measurement approaches, and the establishment of functioning processes and institutions.  

\subsection{Real-world applicability of metrics}

First, it is noteworthy that, historically, evaluation made a difference for safety because it tracked real-world risks. Measuring real-world performance does not mean waiting until risks manifest —- on the contrary, earlier pre-deployment risk detection and evaluation allows for more comprehensive and cheaper mitigations \citep{collingridge_social_1982}. For example, in clinical trials, strict requirements exist for staged, pre-clinical testing in order to minimize risks to vulnerable patient populations. Similarly, airplanes are first designed and tested through simulations to improve understanding of their performance while minimizing risks to life and material damage. 

Pre-deployment testing may help identify real-world risks earlier -- however, it must be accompanied by \textit{post-deployment monitoring} to detect emergent harms as they happen. For instance, unexpected side effects and off-label use of pharmaceuticals in the medical domain, especially on under-tested populations, are nearly impossible to anticipate pre-deployment. Many of these issues only emerge from highly complex interactions at the point of use. In such cases, health providers, patients and manufacturers are required to report adverse events to regulatory agencies via incident databases \footnote{For example: https://open.fda.gov/data/faers/, https://vaers.hhs.gov/, https://yellowcard.mhra.gov.uk/}. The collection of these incidents and the resulting analyses can then be used to inform any restriction or cautionary uses of the drug or vaccine, especially for at-risk populations. As an example, the discovery of Myocarditis symptoms from the COVID-19 vaccine was facilitated by the Vaccine Adverse Events System (VAERS) incident database. This finding led to a warning and an adjusted dosage recommendation for the most impacted population of male vaccine recipients, aged 12 to 17 \citep{oster_myocarditis_2022}. In some cases, monitoring data can even be used to feed back into future pre-deployment evaluation practices -- for example, the results of race-based failures observed in an FDA incident database for medical devices \citep{maude_manufacturer_nodate} informed new health department guidelines on adequate equitable representation in pre-clinical trials for such devices \citep{health_evaluation_2017, fox-rawlings_diversity_2018}.

\subsection{Iteratively Refining Metrics}

The metrics and measurement approaches of evaluation must be iteratively refined and calibrated over time. This iterative process includes choosing and refining relevant measurement \textit{targets}, i.e. the concepts to be measured. Initially, the automotive industry focused on human-caused errors, responding with drivers’ education, drivers’ licenses, and laws against drunk driving. However, as accidents continued to soar, seatbelt regulations and other design choices became a focal point, feeding into notions of a car’s “crashworthiness” tied to manufacturer responsibility \citep{diaz_crashworthiness_2020}. This measurement target of crashworthiness has continued to evolve over time. For example in Europe, concerns about the safety of pedestrians and cyclists were incorporated in an expanded notion of crashworthiness \citep{united_nations_proposal_2011}, broadening what it means for a car to be considered “safe”. 

As a measurement target is refined over time, so are the measurement instruments that are designed to capture it. With the measurement of temperature, divergent thermometer readings revealed the importance of engineering instruments with a reliable liquid indicator \citep{chang_spirit_2001}. Further attempts to calibrate thermometers gave rise to deeper insights about temperature itself -- as an indication of matter phase changes (i.e. Celsius), human body responses (i.e. Fahrenheit), and quantum mechanical properties (i.e. Kelvin). However, no single measurement instrument is perfect -- by triangulating results from multiple methods, more robust insights can be gained \citep{campbell_convergent_1959, jespersen_triangulation_2017}. Ultimately, identifying measurement targets, designing metrics, and developing measurement instruments, are all interdependent tasks that require a careful iterative process. 

\subsection{Establishing Institutions \& Norms}
A successful evaluation ecosystem requires investing in institutions. The advocacy of Harvey Wiley, Samuel Hopkins Adams, and others led to the 1938 passing of the United States Federal Food, Drug, and Cosmetic Act. This act led to the creation of the Food and Drug Administration (FDA), an agency that is now widely known for its rigorous pharmaceutical and nutrition testing regimes. At the FDA, Wiley and his team developed numerous innovative methods for identifying the presence and effects of particular poisonous ingredients, notably leading several multi-year experiments to assess the pernicious effects of various chemicals on a group of volunteers known as the “Poison Squad” \citep{blum_poison_2018}. Without the centralization of testing efforts through a single agency, this team could not have had the resources or coordination capacities to execute such long-term and large-scale experiments. 

In many fields, readily available evaluation tools, shared evaluation infrastructure, and standards afforded by such institutions have contributed to the establishment of more thorough evaluation regimes \citep{vedung_public_2017, timmermans_gold_2003}.  After the number of cars on the road increased by an order of magnitude throughout the early 20th century, the corresponding increase in fatal crashes pushed Ralph Nader and other advocates to establish the National Traffic and Motor Vehicle Safety Act in 1966, responsible for the National Highway Safety Bureau (now the transportation testing agency known as the National Highway Traffic Safety Administration, NHTSA). By 1985, Ralph Nader claimed “programs, which emphasize engineering safety, have saved more than 150,000 lives and prevented or reduced in severity a far larger number of injuries” \citep{nader_opinion_1985}. In 2015, a NHTSA report revealed that this trend has continued, with an estimated 613,501 lives saved between 1960 to 2012 \citep{britannica_samuel_2025}. Nader attributed much of this success to the meaningful enforcement of government-mandated standards, including active monitoring -- i.e., regularly measuring everything from fuel efficiency to auto handling and braking capabilities—by the NHTSA, which led to the recall of millions of defective vehicles and tires by the early 1980s. 

\section{Towards an Evaluation Science for Generative AI}

\subsection{Unique Challenges of Generative AI}
While drawing on lessons from other fields, it is important to understand what makes the challenge of evaluating generative AI systems unique. Other systems -- from personal computers to pharmaceuticals -- can be used for purposes that were not originally intended. However, generative AI systems are often explicitly designed to be open-ended -- that is, underspecified and deliberately versatile in the range of use cases they support \citep{hughes_open-endedness_2024}. This open-endedness makes it hard to define precise measurement targets in AI evaluation, resulting in vague targets such as the long-standing trend of measuring an AI system’s “general intelligence”, rather than performance on specific tasks \citep{raji_ai_2021}. Furthermore, generative AI systems tend to be less \textit{deterministic} -- i.e., the same input can lead to different outputs due to their stochastic nature, and due to unknown factors in training data \citep{raji_bodies_2021}. This non-determinism makes it harder to predict system behaviors compared to prior software systems, as it is difficult to directly trace system design choices -- about training data, model design or the user-interface -- to downstream system outputs and impacts.

Further adding to the complexity of anticipating and evaluating AI system outputs and use cases is the possibility of longitudinal \textit{social} interactions with generative AI systems. This  gives rise to a new class of interaction risks that may evolve in unexpected ways over time (e.g., harmful human–AI “relationships” \citep{manzini_code_2024}. Taken together, these unique challenges inherent to generative AI systems indicate the need for a \textit{behavioral} approach to evaluating such systems, focusing on AI system performance in the context of different real-world settings \citep{rahwan_machine_2019, wagner_measuring_2021, matias_humans_2023}. Indeed, adopting a behavioral approach that treats AI systems as blackboxes can be helpful in enabling some translation between higher-level systemic impact evaluations and lower-level computational methods \citep{mccoy_embers_2024, shiffrin_probing_2023}. 

\subsection{Real-world applicability of metrics}

There is a disconnect between the current AI evaluation culture, with its focus on benchmarking models, and real-world, grounded approaches to the assessment of performance and safety \citep{lazar_ai_2023}. Addressing this divide will require taking deliberate steps to shift the culture surrounding generative AI evaluations from “basic research” toward “\textit{use-inspired} basic research” \citep{stokes_pasteurs_1997}, where the focus is on advancing our scientific understanding of AI system properties and patterns that are relevant for their performance and safety in real-world deployment contexts. 

Evaluations of generative AI systems cannot be “one size fits all.”  As with other fields, even pre-deployment evaluations need to take real-world deployment contexts into account. This echoes several recent calls for holistic, AI system-focused evaluations that take into account relevant context beyond the scope of the current model-focused evaluation culture \citep{lum_bias_2024, goldfarb-tarrant_intrinsic_2021-1, bommasani_trustworthy_2024, saxon_benchmarks_2024, weidinger_sociotechnical_2023}. To achieve this, AI evaluation science must employ a range of approaches that can respond to different evaluation goals, and move beyond coarse grained claims of “general intelligence” towards more task-specific and real-world relevant measures of progress and performance \citep{bowman_what_2021, raji_ai_2021}. A variety of more holistic evaluation methods and instruments, appropriate for differing deployment contexts and evaluation goals, need to be developed \citep{the_national_artificial_intelligence_advisory_committee_naiac_findings_2024, bommasani_trustworthy_2024, solaiman_evaluating_2024, weidinger_sociotechnical_2023, dobbe_system_2022}. By December 2023, less than 6\% of generative AI evaluations accounted for human–AI interactions, and less than 10\% considered broader contextual factors \citep{rauh_gaps_2024}. 

To account for factors beyond technical specifications that influence real-world performance and safety, generative AI evaluations will need to adopt a broader sociotechnical lens \citep{selbst_fairness_2019, chen_sociotechnical_nodate, wallach_evaluating_2024}. Although there is an emerging interest in other approaches, such as more interactive, dynamic, context-rich, and multi-turn benchmarks \citep{chiang_chatbot_2024, saxon_benchmarks_2024, zhou_haicosystem:_2024, magooda_framework_2023}, large gaps remain. For one, anticipating and understanding real-world risks from sustained, personalized human–AI interactions will require more longitudinal studies than have been published to date (e.g., \citet{lai_towards_2023}) and the establishment of post-deployment monitoring regimes for AI systems (e.g., \citet{feng_not_2025}). Furthermore, insights from real-world deployment need to feed back into early-stage evaluation design -- certain existing efforts, such as Anthropic's Clio \citep{anthropic_clio_2024} or AllenAI’s WildBench \citep{lin_wildbench:_2024}, indicate some promise toward an approach of developing pre-deployment benchmarks with data from “naturalistic” interactions from post-deployment contexts. 

\subsection{Iteratively Refining Metrics}

Developing an evaluation science for generative AI systems requires first identifying which concepts should be measured -— that is, to determine the proper measurement  targets. Common targets of interest in the AI context are often abstract and even contested \citep{wallach_evaluating_2024}. Operationally defining metrics that capture these targets involves identifying relevant, tractable subcomponents. Take the widely cited risk of “misinformation”: relevant factors include whether factually correct information is being provided, the subtlety of whether different persons are likely to believe that information, and how such information may be uncritically disseminated. Each of these aspects is best measured at different levels of analysis -- factual accuracy can be determined based on model output, believability requires human-computer interaction studies, and assessing dissemination pathways requires studying the broader systems into which AI is deployed \citep{weidinger_sociotechnical_2023}. Triangulating measurements across these levels of analysis can provide a more holistic picture of “misinformation” propagation. 

Better integration of evaluation metrics across AI development and deployment can be used to further refine, calibrate, and validate these metrics, enabling an iterative scaffolding of this evaluation science \citep{wimsatt_ontology_1994}. Comparing the results of pre-deployment evaluations, such as static benchmarks, to post-deployment evaluations and monitoring enables an evaluation feedback loop, whereby early-stage evaluations can become better calibrated to take real-world deployment contexts into account. For example, comparing results from static benchmark testing and post-deployment monitoring, one might identify that some AI generated computer code is functional, but frequently misunderstood and falsely applied by users. This insight can then be used to improve benchmarks and other early-stage model testing protocols -- e.g. by adopting tests to assess code \textit{legibility}, in addition to testing the functionality of produced computer code \citep{nguyen_how_2024}.

\subsection{Establishing Institutions \& Norms}

A successful evaluation ecosystem requires investment. Current infrastructure falls short of the systematic approach and effectiveness of evaluation regimes in other fields, where evaluation processes are more costly, complex, and distributed between different actors and skill sets \citep{raji_outsider_2022, anthropic_challenges_2023, caliskan_effective_2024}. Prioritizing such investments and developing readily available tools for auditing and evaluation \citep{ojewale_towards_2024} -- including resources to enable the expanded methodological toolkit mentioned above and mechanisms for institutional transparency \citep{whitehouse_removing_2025, caliskan_effective_2024} -- will be critical in order for AI evaluation practice to become systematized, effective and widespread. 

It is already clear that aiming for uncompromised, transparent and open evaluation platforms will come at a significant financial cost. Open source efforts such as Hugging Face’s LLM Leaderboard, Eleutheur AI’s LLM evaluation harness, Stanford’s HELM, and ML Commons provide shared technical infrastructure on which to compare and rank benchmarking results, and there are nascent, but comparable, publicly funded government efforts such as the UK AI Safety Institute’s platform \textit{Inspect} and the US National Institute of Standards and Technology pilot of \textit{ARIA} \footnote{See https://ai-challenges.nist.gov/aria and https://inspect.ai-safety-institute.org.uk/.}. However, running HELM once on the 30 models assessed in 2022 cost USD \$38,000 for the commercial model APIs, and required 20,000 A100 hours of compute to test the open models -- even with Anthropic and Microsoft allowing to run their models for free \citep{liang_holistic_2023}. This differs glaringly from the cost of running an evaluation on traditional benchmarks such as SQuAD \citep{rajpurkar_squad:_2016} or other GLUE tests \citep{wang_glue:_2019}, both of which could be easily downloaded to a personal laptop and run within a few hours at most. Even as specific platforms evolve and expand, this indicates that the next era of evaluation infrastructure for generative AI systems will require financial resources beyond what has been invested so far. Given the history of overlooking the importance of evaluation practices in the machine learning field \citep{paullada_data_2020}, prioritizing and investing in evaluations will be critical to ensuring safe and trustworthy AI systems.

Shared AI evaluation infrastructure can involve much more than just a community leaderboard. Common AI evaluation tools for everything from harm discovery, standard identification and more can facilitate the evaluation process and provide guidance for evaluation best practice across stakeholders in industry and beyond \citep{ojewale_towards_2024, wang2024benchmark}. For instance, many documentation efforts provide direct and indirect guidance to engineering teams on how to approach AI evaluation -- in order to record the requested information in the template, practitioners must, at minimum, satisfy requirements of a particular evaluation process. For instance, the inclusion of disaggregated evaluations in the Model Card template (i.e. evaluating model performance across different demographic subgroups), increased the practice throughout the machine learning field. AI documentation templates such as Model Cards \citep{mitchell_model_2019}, SMACTR \citep{raji_closing_2020}, Datasheets for Datasets, \citep{gebru_datasheets_2021}, and Fact sheets \citep{ibm_using_2024}, as well as multi-year, multi-stakeholder documentation initiatives like ABOUTML \citep{raji_about_2020} all continue to meaningfully guide current model development and evaluation practice -- indeed, several of these documentation templates are being integrated into open-source AI model platforms \citep{liang_whats_2024}, and policy requirements \citep{kawakami_responsible_2024}. New documentation frameworks specific to generative AI evaluation have begun to emerge from corporate alliances between generative AI model developers to advance evaluation norms and standards in this context (e.g. \citet{pais_pais_nodate, frontier_model_forum_frontier_nodate, mlcommons_mlcommons_nodate}. 

\section{Moving Forward}

It is tempting to assume that because generative AI systems are widely used and deployed, they must have been subject to the elaborate safety and performance evaluations that we have come to expect in other fields. Sadly, this is not the case. Because generative AI systems have only recently transitioned from the research landscape to the real world, the current evaluation ecosystem is not yet mature. In many cases, the real-world uses of these systems are still evolving and new application domains are being developed. For many considerations on real-world performance and safety, there are simply no valid, reliable evaluations available yet. Closing this gap requires a deliberate effort to invest in and create an evaluation science for generative AI. 

However, evaluations are not neutral. Choosing what and how to evaluate privileges some issues at the cost of others -- it is not possible to assess all possible use cases and applications, requiring further prioritisation and value judgement. One principled and responsible approach may be to focus on the highest-risk deployment contexts, such as applications in medicine, law, education and finance -- or focus on deployments impacting the most vulnerable populations. A hope may be that by focusing on evaluating generative AI systems safe in these contexts and for these groups, we may lift many boats and build an evaluation ecosystem that makes for more reliable, trustworthy and safe generative AI systems for all.

The trust we have in every product we regularly make use of -- from the toaster used to heat our breakfast, to the vehicle mediating our morning commute -- has been hard-earned. Valuable insights from safety engineering and measurement science in other fields -- such as anticipating real-world failures pre-deployment and monitoring incidents post-deployment, iteratively refining evaluation approaches, and investing in institutions for accessible and robust evaluation ecosystems -- can be adopted to mature practices in the AI field. The unique challenges of generative AI technologies do not absolve the field from this responsibility, but further reinforce a clear need for creating an evaluation science it can call its own.

\section{Acknowledgements}
We thank Sayash Kapoor for comments on this manuscript. 
\typeout{} 
\bibliographystyle{plainnat}
\setcitestyle{maxbibnames=10}
\bibliography{references}

\begin{thebibliography}{106}
\providecommand{\natexlab}[1]{#1}
\providecommand{\url}[1]{\texttt{#1}}
\expandafter\ifx\csname urlstyle\endcsname\relax
  \providecommand{\doi}[1]{doi: #1}\else
  \providecommand{\doi}{doi: \begingroup \urlstyle{rm}\Url}\fi

\bibitem[Anthropic(2023)]{anthropic_challenges_2023}
Anthropic.
\newblock Challenges in evaluating {AI} systems, October 2023.
\newblock URL \url{https://www.anthropic.com/research/evaluating-ai-systems}.

\bibitem[Anthropic(2024{\natexlab{a}})]{anthropic_clio_2024}
Anthropic.
\newblock Clio: {Privacy}-preserving insights into real-world {AI} use,
  December 2024{\natexlab{a}}.
\newblock URL \url{https://www.anthropic.com/research/clio}.

\bibitem[Anthropic(2024{\natexlab{b}})]{anthropic_introducing_2024}
Anthropic.
\newblock Introducing the next generation of {Claude}, March
  2024{\natexlab{b}}.
\newblock URL \url{https://www.anthropic.com/news/claude-3-family}.

\bibitem[Archer and Elliott(2025)]{archer_representation_2025}
Pete Archer and Oli Elliott.
\newblock Representation of {BBC} {News} content in {AI} {Assistants}.
\newblock Technical report, BBC, February 2025.
\newblock URL
  \url{https://www.bbc.co.uk/aboutthebbc/documents/bbc-research-into-ai-assistants.pdf}.

\bibitem[Birhane et~al.(2024)Birhane, Steed, Ojewale, Vecchione, and
  Raji]{birhane_ai_2024}
Abeba Birhane, Ryan Steed, Victor Ojewale, Briana Vecchione, and
  Inioluwa~Deborah Raji.
\newblock {AI} auditing: {The} {Broken} {Bus} on the {Road} to {AI}
  {Accountability}, January 2024.
\newblock URL \url{http://arxiv.org/abs/2401.14462}.
\newblock arXiv:2401.14462.

\bibitem[Blum(2018)]{blum_poison_2018}
Deborah Blum.
\newblock \emph{The {Poison} {Squad}: {One} {Chemist}’s {Single}-{Minded}
  {Crusade} for {Food} {Safety} at the {Turn} of the {Twentieth} {Century}}.
\newblock Penguin, New York, 2018.

\bibitem[Bommasani and Liang(2024)]{bommasani_trustworthy_2024}
Rishi Bommasani and Percy Liang.
\newblock Trustworthy {Social} {Bias} {Measurement}.
\newblock \emph{Proceedings of the AAAI/ACM Conference on AI, Ethics, and
  Society}, 7\penalty0 (1):\penalty0 210--224, October 2024.
\newblock ISSN 3065-8365.
\newblock \doi{10.1609/aies.v7i1.31630}.
\newblock URL \url{https://ojs.aaai.org/index.php/AIES/article/view/31630}.

\bibitem[Bowman and Dahl(2021)]{bowman_what_2021}
Samuel~R. Bowman and George~E. Dahl.
\newblock What {Will} it {Take} to {Fix} {Benchmarking} in {Natural} {Language}
  {Understanding}?, October 2021.
\newblock URL \url{http://arxiv.org/abs/2104.02145}.
\newblock arXiv:2104.02145.

\bibitem[Boyd(2023)]{boyd_microsoft_2023}
Eric Boyd.
\newblock Microsoft and {Epic} expand {AI} collaboration to accelerate
  generative {AI}’s impact in healthcare, addressing the industry’s most
  pressing needs, August 2023.
\newblock URL
  \url{https://blogs.microsoft.com/blog/2023/08/22/microsoft-and-epic-expand-ai-collaboration-to-accelerate-generative-ais-impact-in-healthcare-addressing-the-industrys-most-pressing-needs/}.

\bibitem[Britannica()]{britannica_samuel_2025}
Britannica.
\newblock Samuel hopkins adams {\textbar} biography, works, muckraker, \& facts
  {\textbar} britannica.
\newblock URL \url{https://www.britannica.com/biography/Samuel-Hopkins-Adams}.

\bibitem[Caliskan and Lum(2024)]{caliskan_effective_2024}
Aylin Caliskan and Kristian Lum.
\newblock Effective {AI} regulation requires understanding general-purpose
  {AI}, 2024.
\newblock URL
  \url{https://www.brookings.edu/articles/effective-ai-regulation-requires-understanding-general-purpose-ai/}.

\bibitem[Campbell and Fiske(1959)]{campbell_convergent_1959}
Donald~T. Campbell and Donald~W. Fiske.
\newblock Convergent and {Discriminant} {Validation} by the
  {Multitrait}-{Multimethod} {Matrix}.
\newblock \emph{Psychological Bulletin}, 56\penalty0 (2):\penalty0 81--105,
  1959.
\newblock \doi{https://doi.org/10.1037/h0046016}.

\bibitem[Chang(2001)]{chang_spirit_2001}
Hasok Chang.
\newblock Spirit, air, and quicksilver: {The} search for the "real" scale of
  temperature.
\newblock \emph{Historical Studies in the Physical and Biological Sciences},
  31\penalty0 (2):\penalty0 249--284, March 2001.
\newblock ISSN 0890-9997.
\newblock \doi{10.1525/hsps.2001.31.2.249}.
\newblock URL
  \url{https://online.ucpress.edu/hsns/article/31/2/249/47889/Spirit-air-and-quicksilver-The-search-for-the-real}.

\bibitem[Chen and Metcalf()]{chen_sociotechnical_nodate}
Brian~J. Chen and Jacob Metcalf.
\newblock A {Sociotechnical} {Approach} to {AI} {Policy}.
\newblock URL
  \url{https://datasociety.net/library/a-sociotechnical-approach-to-ai-policy/}.

\bibitem[Chiang et~al.(2024)Chiang, Zheng, Sheng, Angelopoulos, Li, Li, Zhang,
  Zhu, Jordan, Gonzalez, and Stoica]{chiang_chatbot_2024}
Wei-Lin Chiang, Lianmin Zheng, Ying Sheng, Anastasios~Nikolas Angelopoulos,
  Tianle Li, Dacheng Li, Hao Zhang, Banghua Zhu, Michael Jordan, Joseph~E.
  Gonzalez, and Ion Stoica.
\newblock Chatbot {Arena}: {An} {Open} {Platform} for {Evaluating} {LLMs} by
  {Human} {Preference}, March 2024.
\newblock URL \url{http://arxiv.org/abs/2403.04132}.
\newblock arXiv:2403.04132.

\bibitem[Collingridge(1982)]{collingridge_social_1982}
David Collingridge.
\newblock \emph{The {Social} {Control} of {Technology}}.
\newblock Pinter Publishers, London, 1982.

\bibitem[de~Vries et~al.(2020)de~Vries, Bahdanau, and
  Manning]{de_vries_towards_2020}
Harm de~Vries, Dzmitry Bahdanau, and Christopher Manning.
\newblock Towards {Ecologically} {Valid} {Research} on {Language} {User}
  {Interfaces}, July 2020.
\newblock URL \url{http://arxiv.org/abs/2007.14435}.
\newblock arXiv:2007.14435.

\bibitem[Dobbe(2022)]{dobbe_system_2022}
Roel I.~J. Dobbe.
\newblock System {Safety} and {Artificial} {Intelligence}, February 2022.
\newblock URL \url{http://arxiv.org/abs/2202.09292}.
\newblock arXiv:2202.09292.

\bibitem[Dzieza(2024)]{dzieza_confusing_2024}
Josh Dzieza.
\newblock The confusing reality of {AI} friends, December 2024.
\newblock URL
  \url{https://www.theverge.com/c/24300623/ai-companions-replika-openai-chatgpt-assistant-romance}.

\bibitem[Díaz and Costas(2020)]{diaz_crashworthiness_2020}
Jacobo Díaz and Miguel Costas.
\newblock Crashworthiness.
\newblock In Holm Altenbach and Andreas Öchsner, editors, \emph{Encyclopedia
  of {Continuum} {Mechanics}}, pages 469--486. Springer, Berlin, Heidelberg,
  2020.
\newblock ISBN 9783662557716.
\newblock \doi{10.1007/978-3-662-55771-6_223}.
\newblock URL \url{https://doi.org/10.1007/978-3-662-55771-6_223}.

\bibitem[{European Commission}(2024)]{EC_second_2024}
{European Commission}.
\newblock Second {Draft} of the {General}-{Purpose} {AI} {Code} of {Practice}
  published, written by independent experts, December 2024.
\newblock URL
  \url{https://digital-strategy.ec.europa.eu/en/library/second-draft-general-purpose-ai-code-practice-published-written-independent-experts}.

\bibitem[Faverio and Tyson(2023)]{faverio_what_2023}
Michelle Faverio and Alec Tyson.
\newblock What the data says about {Americans}’ views of artificial
  intelligence, November 2023.
\newblock URL
  \url{https://www.pewresearch.org/short-reads/2023/11/21/what-the-data-says-about-americans-views-of-artificial-intelligence/}.

\bibitem[{FDA MAUDE Database}()]{maude_manufacturer_nodate}
{FDA MAUDE Database}.
\newblock Manufacturer and {User} {Facility} {Device} {Experience} ({MAUDE})
  {Database}.
\newblock URL
  \url{https://www.accessdata.fda.gov/scripts/cdrh/cfdocs/cfmaude/search.cfm}.

\bibitem[Feng et~al.(2025)Feng, Xia, Singh, and Pirracchio]{feng_not_2025}
Jean Feng, Fan Xia, Karandeep Singh, and Romain Pirracchio.
\newblock Not {All} {Clinical} {AI} {Monitoring} {Systems} {Are} {Created}
  {Equal}: {Review} and {Recommendations}.
\newblock \emph{NEJM AI}, 2\penalty0 (2), January 2025.
\newblock ISSN 2836-9386.
\newblock \doi{10.1056/AIra2400657}.
\newblock URL \url{https://ai.nejm.org/doi/10.1056/AIra2400657}.

\bibitem[Fox-Rawlings et~al.(2018)Fox-Rawlings, Gottschalk, Doamekpor, and
  Zuckerman]{fox-rawlings_diversity_2018}
Stephanie~R. Fox-Rawlings, Laura~B. Gottschalk, Laurén~A. Doamekpor, and
  Diana~M. Zuckerman.
\newblock Diversity in {Medical} {Device} {Clinical} {Trials}: {Do} {We} {Know}
  {What} {Works} for {Which} {Patients}?
\newblock \emph{The Milbank Quarterly}, 96\penalty0 (3):\penalty0 499--529,
  September 2018.
\newblock ISSN 0887-378X.
\newblock \doi{10.1111/1468-0009.12344}.
\newblock URL \url{https://www.ncbi.nlm.nih.gov/pmc/articles/PMC6131322/}.

\bibitem[Friedler et~al.(2023)Friedler, Singh, Blili-Hamelin, Metcalfe, and
  Chen]{friedler_ai_2023}
Sorelle Friedler, Ranjit Singh, Borhane Blili-Hamelin, Jacob Metcalfe, and
  Brian~J. Chen.
\newblock {AI} {Red}-{Teaming} {Is} {Not} a {One}-{Stop} {Solution} to {AI}
  {Harms}: {Recommendations} for {Using} {Red}-{Teaming} for {AI}
  {Accountability}, 2023.
\newblock URL
  \url{https://datasociety.net/library/ai-red-teaming-is-not-a-one-stop-solution-to-ai-harms-recommendations-for-using-red-teaming-for-ai-accountability/}.

\bibitem[{Frontier Model Forum}()]{frontier_model_forum_frontier_nodate}
{Frontier Model Forum}.
\newblock Frontier {Model} {Forum}: advancing frontier {AI} safety and
  security.
\newblock URL \url{https://www.frontiermodelforum.org/}.

\bibitem[Gebru et~al.(2021)Gebru, Morgenstern, Vecchione, Vaughan, Wallach,
  III, and Crawford]{gebru_datasheets_2021}
Timnit Gebru, Jamie Morgenstern, Briana Vecchione, Jennifer~Wortman Vaughan,
  Hanna Wallach, Hal~Daumé III, and Kate Crawford.
\newblock Datasheets for {Datasets}, December 2021.
\newblock URL \url{http://arxiv.org/abs/1803.09010}.
\newblock arXiv:1803.09010.

\bibitem[Gemini et~al.(2024)Gemini, Georgiev, Lei, Burnell, Bai, Gulati,
  Tanzer, Vincent, Pan, Wang, Mariooryad, Ding, Geng, Alcober, Frostig,
  Omernick, Walker, Paduraru, Sorokin, Tacchetti, Gaffney, Daruki, Sercinoglu,
  Gleicher, Love, Voigtlaender, Jain, Surita, Mohamed, Blevins, Ahn, Zhu,
  Kawintiranon, Firat, Gu, Zhang, Rahtz, Faruqui, Clay, Gilmer, Co-Reyes,
  Penchev, Zhu, Morioka, Hui, Haridasan, Campos, Mahdieh, Guo, and Hassan~et
  al.]{gemini_team_gemini_2024}
Gemini, Petko Georgiev, Ving~Ian Lei, Ryan Burnell, Libin Bai, Anmol Gulati,
  Garrett Tanzer, Damien Vincent, Zhufeng Pan, Shibo Wang, Soroosh Mariooryad,
  Yifan Ding, Xinyang Geng, Fred Alcober, Roy Frostig, Mark Omernick, Lexi
  Walker, Cosmin Paduraru, Christina Sorokin, Andrea Tacchetti, Colin Gaffney,
  Samira Daruki, Olcan Sercinoglu, Zach Gleicher, Juliette Love, Paul
  Voigtlaender, Rohan Jain, Gabriela Surita, Kareem Mohamed, Rory Blevins,
  Junwhan Ahn, Tao Zhu, Kornraphop Kawintiranon, Orhan Firat, Yiming Gu, Yujing
  Zhang, Matthew Rahtz, Manaal Faruqui, Natalie Clay, Justin Gilmer, J.~D.
  Co-Reyes, Ivo Penchev, Rui Zhu, Nobuyuki Morioka, Kevin Hui, Krishna
  Haridasan, Victor Campos, Mahdis Mahdieh, Mandy Guo, and Samer Hassan~et al.
\newblock Gemini 1.5: {Unlocking} multimodal understanding across millions of
  tokens of context, December 2024.
\newblock URL \url{http://arxiv.org/abs/2403.05530}.
\newblock arXiv:2403.05530.

\bibitem[Goldfarb-Tarrant et~al.(2021{\natexlab{a}})Goldfarb-Tarrant, Marchant,
  Sanchez, Pandya, and Lopez]{goldfarb-tarrant_intrinsic_2021}
Seraphina Goldfarb-Tarrant, Rebecca Marchant, Ricardo~Muñoz Sanchez, Mugdha
  Pandya, and Adam Lopez.
\newblock Intrinsic {Bias} {Metrics} {Do} {Not} {Correlate} with {Application}
  {Bias}, June 2021{\natexlab{a}}.
\newblock URL \url{http://arxiv.org/abs/2012.15859}.
\newblock arXiv:2012.15859.

\bibitem[Goldfarb-Tarrant et~al.(2021{\natexlab{b}})Goldfarb-Tarrant, Marchant,
  Sanchez, Pandya, and Lopez]{goldfarb-tarrant_intrinsic_2021-1}
Seraphina Goldfarb-Tarrant, Rebecca Marchant, Ricardo~Muñoz Sanchez, Mugdha
  Pandya, and Adam Lopez.
\newblock Intrinsic {Bias} {Metrics} {Do} {Not} {Correlate} with {Application}
  {Bias}, June 2021{\natexlab{b}}.
\newblock URL \url{http://arxiv.org/abs/2012.15859}.
\newblock arXiv:2012.15859.

\bibitem[Grattafiori et~al.(2024)Grattafiori, Dubey, Jauhri, Pandey, Kadian,
  Al-Dahle, Letman, Mathur, Schelten, Vaughan, Yang, Fan, Goyal, Hartshorn,
  Yang, Mitra, Sravankumar, Korenev, Hinsvark, Rao, Zhang, Rodriguez,
  Gregerson, Spataru, Roziere, Biron, Tang, Chern, Caucheteux, Nayak, Bi,
  Marra, McConnell, Keller, Touret, Wu, Wong, Ferrer, Nikolaidis, Allonsius,
  Song, Pintz, Livshits, Wyatt, Esiobu, Choudhary, Mahajan, Garcia-Olano,
  Perino, Hupkes, Lakomkin, and AlBadawy~et al.]{grattafiori_llama_2024}
Aaron Grattafiori, Abhimanyu Dubey, Abhinav Jauhri, Abhinav Pandey, Abhishek
  Kadian, Ahmad Al-Dahle, Aiesha Letman, Akhil Mathur, Alan Schelten, Alex
  Vaughan, Amy Yang, Angela Fan, Anirudh Goyal, Anthony Hartshorn, Aobo Yang,
  Archi Mitra, Archie Sravankumar, Artem Korenev, Arthur Hinsvark, Arun Rao,
  Aston Zhang, Aurelien Rodriguez, Austen Gregerson, Ava Spataru, Baptiste
  Roziere, Bethany Biron, Binh Tang, Bobbie Chern, Charlotte Caucheteux, Chaya
  Nayak, Chloe Bi, Chris Marra, Chris McConnell, Christian Keller, Christophe
  Touret, Chunyang Wu, Corinne Wong, Cristian~Canton Ferrer, Cyrus Nikolaidis,
  Damien Allonsius, Daniel Song, Danielle Pintz, Danny Livshits, Danny Wyatt,
  David Esiobu, Dhruv Choudhary, Dhruv Mahajan, Diego Garcia-Olano, Diego
  Perino, Dieuwke Hupkes, Egor Lakomkin, and Ehab AlBadawy~et al.
\newblock The {Llama} 3 {Herd} of {Models}, November 2024.
\newblock URL \url{http://arxiv.org/abs/2407.21783}.
\newblock arXiv:2407.21783.

\bibitem[Heaven(2022)]{heaven_why_2022}
Will~Douglas Heaven.
\newblock Why {Meta}’s latest large language model survived only three days
  online, November 2022.
\newblock URL
  \url{https://www.technologyreview.com/2022/11/18/1063487/meta-large-language-model-ai-only-survived-three-days-gpt-3-science/}.

\bibitem[Hu(2023)]{hu_chatgpt_2023}
Krystal Hu.
\newblock {ChatGPT} sets record for fastest-growing user base - analyst note,
  February 2023.
\newblock URL
  \url{https://www.reuters.com/technology/chatgpt-sets-record-fastest-growing-user-base-analyst-note-2023-02-01/}.

\bibitem[Hughes et~al.(2024)Hughes, Dennis, Parker-Holder, Behbahani,
  Mavalankar, Shi, Schaul, and Rocktaschel]{hughes_open-endedness_2024}
Edward Hughes, Michael Dennis, Jack Parker-Holder, Feryal Behbahani, Aditi
  Mavalankar, Yuge Shi, Tom Schaul, and Tim Rocktaschel.
\newblock Open-{Endedness} is {Essential} for {Artificial} {Superhuman}
  {Intelligence}, June 2024.
\newblock URL \url{http://arxiv.org/abs/2406.04268}.
\newblock arXiv:2406.04268.

\bibitem[IBM(year)]{ibm_using_2024}
IBM.
\newblock Using {AI} {Factsheets} for {AI} {Governance} {\textbar} {IBM}
  {Cloud} {Pak} for {Data} as a {Service}, year.
\newblock URL
  \url{https://dataplatform.cloud.ibm.com/docs/content/wsj/analyze-data/factsheets-model-inventory.html?context=cpdaas}.

\bibitem[Jespersen and Wallace(2017)]{jespersen_triangulation_2017}
Lone Jespersen and Carol~A. Wallace.
\newblock Triangulation and the importance of establishing valid methods for
  food safety culture evaluation.
\newblock \emph{Food Research International}, 100:\penalty0 244--253, October
  2017.
\newblock ISSN 0963-9969.
\newblock \doi{10.1016/j.foodres.2017.07.009}.
\newblock URL
  \url{https://www.sciencedirect.com/science/article/pii/S0963996917303319}.

\bibitem[Johnson et~al.(2025)Johnson, Silva, Leon, Eslami, Schwanke, Dotan, and
  Heidari]{johnson_legacy_2025}
Nari Johnson, Elise Silva, Harrison Leon, Motahhare Eslami, Beth Schwanke,
  Ravit Dotan, and Hoda Heidari.
\newblock Legacy {Procurement} {Practices} {Shape} {How} {U}.{S}. {Cities}
  {Govern} {AI}: {Understanding} {Government} {Employees}' {Practices},
  {Challenges}, and {Needs}, February 2025.
\newblock URL \url{http://arxiv.org/abs/2411.04994}.
\newblock arXiv:2411.04994.

\bibitem[Kawakami et~al.(2024)Kawakami, Wilkinson, and
  Chouldechova]{kawakami_responsible_2024}
Anna Kawakami, Daricia Wilkinson, and Alexandra Chouldechova.
\newblock Do {Responsible} {AI} {Artifacts} {Advance} {Stakeholder} {Goals}?
  {Four} {Key} {Barriers} {Perceived} by {Legal} and {Civil} {Stakeholders},
  August 2024.
\newblock URL \url{http://arxiv.org/abs/2408.12047}.
\newblock arXiv:2408.12047.

\bibitem[Kiela et~al.(2021)Kiela, Bartolo, Nie, Kaushik, Geiger, Wu, Vidgen,
  Prasad, Singh, Ringshia, Ma, Thrush, Riedel, Waseem, Stenetorp, Jia, Bansal,
  Potts, and Williams]{kiela_dynabench:_2021}
Douwe Kiela, Max Bartolo, Yixin Nie, Divyansh Kaushik, Atticus Geiger,
  Zhengxuan Wu, Bertie Vidgen, Grusha Prasad, Amanpreet Singh, Pratik Ringshia,
  Zhiyi Ma, Tristan Thrush, Sebastian Riedel, Zeerak Waseem, Pontus Stenetorp,
  Robin Jia, Mohit Bansal, Christopher Potts, and Adina Williams.
\newblock Dynabench: {Rethinking} {Benchmarking} in {NLP}, April 2021.
\newblock URL \url{http://arxiv.org/abs/2104.14337}.
\newblock arXiv:2104.14337.

\bibitem[Lai et~al.(2023)Lai, Chen, Smith-Renner, Liao, and
  Tan]{lai_towards_2023}
Vivian Lai, Chacha Chen, Alison Smith-Renner, Q.~Vera Liao, and Chenhao Tan.
\newblock Towards a {Science} of {Human}-{AI} {Decision} {Making}: {An}
  {Overview} of {Design} {Space} in {Empirical} {Human}-{Subject} {Studies}.
\newblock In \emph{Proceedings of the 2023 {ACM} {Conference} on {Fairness},
  {Accountability}, and {Transparency}}, {FAccT} '23, pages 1369--1385, New
  York, NY, USA, June 2023. Association for Computing Machinery.
\newblock ISBN 9798400701924.
\newblock \doi{10.1145/3593013.3594087}.
\newblock URL \url{https://dl.acm.org/doi/10.1145/3593013.3594087}.

\bibitem[Lazar and Nelson(2023)]{lazar_ai_2023}
Seth Lazar and Alondra Nelson.
\newblock {AI} safety on whose terms?
\newblock \emph{Science}, 381\penalty0 (6654):\penalty0 138--138, July 2023.
\newblock ISSN 0036-8075, 1095-9203.
\newblock \doi{10.1126/science.adi8982}.
\newblock URL \url{https://www.science.org/doi/10.1126/science.adi8982}.

\bibitem[Liang et~al.(2023)Liang, Bommasani, Lee, Tsipras, Soylu, Yasunaga,
  Zhang, Narayanan, Wu, Kumar, Newman, Yuan, Yan, Zhang, Cosgrove, Manning,
  Ré, Acosta-Navas, Hudson, Zelikman, Durmus, Ladhak, Rong, Ren, Yao, Wang,
  Santhanam, Orr, Zheng, Yuksekgonul, Suzgun, Kim, Guha, Chatterji, Khattab,
  Henderson, Huang, Chi, Xie, Santurkar, Ganguli, Hashimoto, Icard, Zhang,
  Chaudhary, Wang, Li, Mai, Zhang, and Koreeda]{liang_holistic_2023}
Percy Liang, Rishi Bommasani, Tony Lee, Dimitris Tsipras, Dilara Soylu,
  Michihiro Yasunaga, Yian Zhang, Deepak Narayanan, Yuhuai Wu, Ananya Kumar,
  Benjamin Newman, Binhang Yuan, Bobby Yan, Ce~Zhang, Christian Cosgrove,
  Christopher~D. Manning, Christopher Ré, Diana Acosta-Navas, Drew~A. Hudson,
  Eric Zelikman, Esin Durmus, Faisal Ladhak, Frieda Rong, Hongyu Ren, Huaxiu
  Yao, Jue Wang, Keshav Santhanam, Laurel Orr, Lucia Zheng, Mert Yuksekgonul,
  Mirac Suzgun, Nathan Kim, Neel Guha, Niladri Chatterji, Omar Khattab, Peter
  Henderson, Qian Huang, Ryan Chi, Sang~Michael Xie, Shibani Santurkar, Surya
  Ganguli, Tatsunori Hashimoto, Thomas Icard, Tianyi Zhang, Vishrav Chaudhary,
  William Wang, Xuechen Li, Yifan Mai, Yuhui Zhang, and Yuta Koreeda.
\newblock Holistic {Evaluation} of {Language} {Models}, October 2023.
\newblock URL \url{http://arxiv.org/abs/2211.09110}.
\newblock arXiv:2211.09110.

\bibitem[Liang et~al.(2024)Liang, Rajani, Yang, Ozoani, Wu, Chen, Smith, and
  Zou]{liang_whats_2024}
Weixin Liang, Nazneen Rajani, Xinyu Yang, Ezinwanne Ozoani, Eric Wu, Yiqun
  Chen, Daniel~Scott Smith, and James Zou.
\newblock What's documented in {AI}? {Systematic} {Analysis} of {32K} {AI}
  {Model} {Cards}, February 2024.
\newblock URL \url{http://arxiv.org/abs/2402.05160}.
\newblock arXiv:2402.05160.

\bibitem[Liao et~al.(2021)Liao, Taori, Raji, and Schmidt]{liao_are_2021}
Thomas Liao, Rohan Taori, Inioluwa~Deborah Raji, and Ludwig Schmidt.
\newblock Are {We} {Learning} {Yet}? {A} {Meta} {Review} of {Evaluation}
  {Failures} {Across} {Machine} {Learning}.
\newblock August 2021.
\newblock URL \url{https://openreview.net/forum?id=mPducS1MsEK}.

\bibitem[Lin et~al.(2024)Lin, Deng, Chandu, Brahman, Ravichander, Pyatkin,
  Dziri, Bras, and Choi]{lin_wildbench:_2024}
Bill~Yuchen Lin, Yuntian Deng, Khyathi Chandu, Faeze Brahman, Abhilasha
  Ravichander, Valentina Pyatkin, Nouha Dziri, Ronan~Le Bras, and Yejin Choi.
\newblock {WildBench}: {Benchmarking} {LLMs} with {Challenging} {Tasks} from
  {Real} {Users} in the {Wild}, October 2024.
\newblock URL \url{http://arxiv.org/abs/2406.04770}.
\newblock arXiv:2406.04770.

\bibitem[Lum et~al.(2024)Lum, Anthis, Nagpal, and D'Amour]{lum_bias_2024}
Kristian Lum, Jacy~Reese Anthis, Chirag Nagpal, and Alexander D'Amour.
\newblock Bias in {Language} {Models}: {Beyond} {Trick} {Tests} and {Toward}
  {RUTEd} {Evaluation}, February 2024.
\newblock URL \url{http://arxiv.org/abs/2402.12649}.
\newblock arXiv:2402.12649 version: 1.

\bibitem[Magesh et~al.(2024)Magesh, Surani, Dahl, Suzgun, Manning, and
  Ho]{magesh_ai_2024}
Varun Magesh, Faiz Surani, Matthew Dahl, Mirac Suzgun, Christopher~D. Manning,
  and Daniel~E. Ho.
\newblock {AI} on trial: legal models hallucinate in 1 out of 6 (or more)
  benchmarking queries, May 2024.
\newblock URL
  \url{https://hai.stanford.edu/news/ai-trial-legal-models-hallucinate-1-out-6-or-more-benchmarking-queries}.

\bibitem[Magooda et~al.(2023)Magooda, Helyar, Jackson, Sullivan, Atalla, Sheng,
  Vann, Edgar, Palangi, Lutz, Kong, Yun, Kamal, Zarfati, Wallach, Bird, and
  Chen]{magooda_framework_2023}
Ahmed Magooda, Alec Helyar, Kyle Jackson, David Sullivan, Chad Atalla, Emily
  Sheng, Dan Vann, Richard Edgar, Hamid Palangi, Roman Lutz, Hongliang Kong,
  Vincent Yun, Eslam Kamal, Federico Zarfati, Hanna Wallach, Sarah Bird, and
  Mei Chen.
\newblock A {Framework} for {Automated} {Measurement} of {Responsible} {AI}
  {Harms} in {Generative} {AI} {Applications}, October 2023.
\newblock URL \url{http://arxiv.org/abs/2310.17750}.
\newblock arXiv:2310.17750.

\bibitem[Manzini et~al.(2024)Manzini, Keeling, Alberts, Vallor, Morris, and
  Gabriel]{manzini_code_2024}
Arianna Manzini, Geoff Keeling, Lize Alberts, Shannon Vallor, Meredith~Ringel
  Morris, and Iason Gabriel.
\newblock The {Code} {That} {Binds} {Us}: {Navigating} the {Appropriateness} of
  {Human}-{AI} {Assistant} {Relationships}.
\newblock \emph{Proceedings of the AAAI/ACM Conference on AI, Ethics, and
  Society}, 7\penalty0 (1):\penalty0 943--957, October 2024.
\newblock ISSN 3065-8365.
\newblock \doi{10.1609/aies.v7i1.31694}.
\newblock URL \url{https://ojs.aaai.org/index.php/AIES/article/view/31694}.

\bibitem[Matias(2023)]{matias_humans_2023}
J.~Nathan Matias.
\newblock Humans and algorithms work together — so study them together.
\newblock \emph{Nature}, 617\penalty0 (7960):\penalty0 248--251, May 2023.
\newblock \doi{10.1038/d41586-023-01521-z}.
\newblock URL \url{https://www.nature.com/articles/d41586-023-01521-z}.

\bibitem[McCoy et~al.(2024)McCoy, Yao, Friedman, Hardy, and
  Griffiths]{mccoy_embers_2024}
R.~Thomas McCoy, Shunyu Yao, Dan Friedman, Mathew~D. Hardy, and Thomas~L.
  Griffiths.
\newblock Embers of autoregression show how large language models are shaped by
  the problem they are trained to solve.
\newblock \emph{Proceedings of the National Academy of Sciences}, 121\penalty0
  (41):\penalty0 e2322420121, October 2024.
\newblock ISSN 0027-8424, 1091-6490.
\newblock \doi{10.1073/pnas.2322420121}.
\newblock URL \url{https://pnas.org/doi/10.1073/pnas.2322420121}.

\bibitem[Mirzadeh et~al.(2024)Mirzadeh, Alizadeh, Shahrokhi, Tuzel, Bengio, and
  Farajtabar]{mirzadeh_gsm-symbolic:_2024}
Iman Mirzadeh, Keivan Alizadeh, Hooman Shahrokhi, Oncel Tuzel, Samy Bengio, and
  Mehrdad Farajtabar.
\newblock {GSM}-{Symbolic}: {Understanding} the {Limitations} of {Mathematical}
  {Reasoning} in {Large} {Language} {Models}, October 2024.
\newblock URL \url{http://arxiv.org/abs/2410.05229}.
\newblock arXiv:2410.05229.

\bibitem[Mitchell et~al.(2019)Mitchell, Wu, Zaldivar, Barnes, Vasserman,
  Hutchinson, Spitzer, Raji, and Gebru]{mitchell_model_2019}
Margaret Mitchell, Simone Wu, Andrew Zaldivar, Parker Barnes, Lucy Vasserman,
  Ben Hutchinson, Elena Spitzer, Inioluwa~Deborah Raji, and Timnit Gebru.
\newblock Model {Cards} for {Model} {Reporting}, January 2019.
\newblock URL \url{http://arxiv.org/abs/1810.03993}.
\newblock arXiv:1810.03993.

\bibitem[MLCommons()]{mlcommons_mlcommons_nodate}
MLCommons.
\newblock {MLCommons} - {Better} {AI} for {Everyone}.
\newblock URL \url{https://mlcommons.org/}.

\bibitem[Mulligan()]{mulligan_way_nodate}
Scott~J. Mulligan.
\newblock The way we measure progress in {AI} is terrible.
\newblock URL
  \url{https://www.technologyreview.com/2024/11/26/1107346/the-way-we-measure-progress-in-ai-is-terrible/}.

\bibitem[Murphy~Kelly(2023)]{murphy_kelly_microsofts_2023}
Samantha Murphy~Kelly.
\newblock Microsoft’s {Bing} {AI} demo called out for several errors
  {\textbar} {CNN} {Business}, February 2023.
\newblock URL
  \url{https://www.cnn.com/2023/02/14/tech/microsoft-bing-ai-errors/index.html}.

\bibitem[Nader(1985)]{nader_opinion_1985}
Ralph Nader.
\newblock Opinion {\textbar} {How} law has improved auto technology.
\newblock \emph{The New York Times}, December 1985.
\newblock ISSN 0362-4331.
\newblock URL
  \url{https://www.nytimes.com/1985/12/29/opinion/how-law-has-improved-auto-technology.html}.

\bibitem[{NAIAC}(2024)]{the_national_artificial_intelligence_advisory_committee_naiac_findings_2024}
{The National Artificial Intelligence Advisory Committee} {NAIAC}.
\newblock Findings \& recommendations: {AI} safety.
\newblock Technical report, May 2024.
\newblock URL
  \url{https://ai.gov/wp-content/uploads/2024/06/FINDINGS-RECOMMENDATIONS_AI-Safety.pdf}.

\bibitem[{National Institute of Standards {and}
  Technology}(2021)]{NIST_ai_2021}
{National Institute of Standards {and} Technology}.
\newblock {AI} {Risk} {Management} {Framework}, July 2021.
\newblock URL \url{https://www.nist.gov/itl/ai-risk-management-framework}.

\bibitem[Nguyen et~al.(2024)Nguyen, McLean~Babe, Zi, Guha, Anderson, and
  Feldman]{nguyen_how_2024}
Sydney Nguyen, Hannah McLean~Babe, Yangtian Zi, Arjun Guha, Carolyn~Jane
  Anderson, and Molly~Q Feldman.
\newblock How {Beginning} {Programmers} and {Code} {LLMs} ({Mis})read {Each}
  {Other}, 2024.
\newblock URL \url{https://arxiv.org/html/2401.15232v1}.

\bibitem[Nicholson(2023)]{nicholson_bing_2023}
Katie Nicholson.
\newblock Bing chatbot says it feels violated and exposed after attack.
\newblock \emph{CBC News}, February 2023.
\newblock URL
  \url{https://www.cbc.ca/news/science/bing-chatbot-ai-hack-1.6752490}.

\bibitem[Ojewale et~al.(2024)Ojewale, Steed, Vecchione, Birhane, and
  Raji]{ojewale_towards_2024}
Victor Ojewale, Ryan Steed, Briana Vecchione, Abeba Birhane, and
  Inioluwa~Deborah Raji.
\newblock Towards {AI} {Accountability} {Infrastructure}: {Gaps} and
  {Opportunities} in {AI} {Audit} {Tooling}, March 2024.
\newblock URL \url{http://arxiv.org/abs/2402.17861}.
\newblock arXiv:2402.17861.

\bibitem[Omiye et~al.(2023)Omiye, Lester, Spichak, Rotemberg, and
  Daneshjou]{omiye_large_2023}
Jesutofunmi~A. Omiye, Jenna~C. Lester, Simon Spichak, Veronica Rotemberg, and
  Roxana Daneshjou.
\newblock Large language models propagate race-based medicine.
\newblock \emph{npj Digital Medicine}, 6\penalty0 (1):\penalty0 1--4, October
  2023.
\newblock ISSN 2398-6352.
\newblock \doi{10.1038/s41746-023-00939-z}.
\newblock URL \url{https://www.nature.com/articles/s41746-023-00939-z}.

\bibitem[OpenAI(2024)]{noauthor_gpt-4o_nodate}
OpenAI.
\newblock {GPT}-4o {System} {Card}, 2024.
\newblock URL \url{https://openai.com/index/gpt-4o-system-card/}.
\newblock publisher:.

\bibitem[Oster et~al.(2022)Oster, Shay, Su, Gee, Creech, Broder, Edwards,
  Soslow, Dendy, Schlaudecker, Lang, Barnett, Ruberg, Smith, Campbell, Lopes,
  Sperling, Baumblatt, Thompson, Marquez, Strid, Woo, Pugsley, Reagan-Steiner,
  DeStefano, and Shimabukuro]{oster_myocarditis_2022}
Matthew~E. Oster, David~K. Shay, John~R. Su, Julianne Gee, C.~Buddy Creech,
  Karen~R. Broder, Kathryn Edwards, Jonathan~H. Soslow, Jeffrey~M. Dendy,
  Elizabeth Schlaudecker, Sean~M. Lang, Elizabeth~D. Barnett, Frederick~L.
  Ruberg, Michael~J. Smith, M.~Jay Campbell, Renato~D. Lopes, Laurence~S.
  Sperling, Jane~A. Baumblatt, Deborah~L. Thompson, Paige~L. Marquez, Penelope
  Strid, Jared Woo, River Pugsley, Sarah Reagan-Steiner, Frank DeStefano, and
  Tom~T. Shimabukuro.
\newblock Myocarditis cases reported after {mRNA}-based {COVID}-19 vaccination
  in the {US} from {December} 2020 to {August} 2021.
\newblock \emph{JAMA}, 327\penalty0 (4):\penalty0 331--40, January 2022.
\newblock URL \url{https://jamanetwork.com/journals/jama/fullarticle/2788346}.

\bibitem[{Partnership on AI}()]{pais_pais_nodate}
{Partnership on AI}.
\newblock {PAI}’s {Guidance} for {Safe} {Foundation} {Model} {Deployment}.
\newblock URL \url{https://partnershiponai.org/modeldeployment/}.

\bibitem[Paullada et~al.(2020)Paullada, Raji, Bender, Denton, and
  Hanna]{paullada_data_2020}
Amandalynne Paullada, Inioluwa~Deborah Raji, Emily~M. Bender, Emily Denton, and
  Alex Hanna.
\newblock Data and its (dis)contents: {A} survey of dataset development and use
  in machine learning research, December 2020.
\newblock URL \url{http://arxiv.org/abs/2012.05345}.
\newblock arXiv:2012.05345.

\bibitem[Rahwan et~al.(2019)Rahwan, Cebrian, Obradovich, Bongard, Bonnefon,
  Breazeal, Crandall, Christakis, Couzin, Jackson, Jennings, Kamar, Kloumann,
  Larochelle, Lazer, McElreath, Mislove, Parkes, Pentland, Roberts, Shariff,
  Tenenbaum, and Wellman]{rahwan_machine_2019}
Iyad Rahwan, Manuel Cebrian, Nick Obradovich, Josh Bongard, Jean-François
  Bonnefon, Cynthia Breazeal, Jacob~W. Crandall, Nicholas~A. Christakis,
  Iain~D. Couzin, Matthew~O. Jackson, Nicholas~R. Jennings, Ece Kamar,
  Isabel~M. Kloumann, Hugo Larochelle, David Lazer, Richard McElreath, Alan
  Mislove, David~C. Parkes, Alex~‘Sandy’ Pentland, Margaret~E. Roberts,
  Azim Shariff, Joshua~B. Tenenbaum, and Michael Wellman.
\newblock Machine behaviour.
\newblock \emph{Nature}, 568\penalty0 (7753):\penalty0 477--486, April 2019.
\newblock ISSN 1476-4687.
\newblock \doi{10.1038/s41586-019-1138-y}.
\newblock URL \url{https://www.nature.com/articles/s41586-019-1138-y}.

\bibitem[Raji(2021)]{raji_bodies_2021}
Deborah Raji.
\newblock The bodies underneath the rubble.
\newblock In Frederike Kaltheuner, editor, \emph{Fake {AI}}. Meatspace Press,
  2021.

\bibitem[Raji and Dobbe(2023)]{raji_concrete_2023}
Inioluwa~Deborah Raji and Roel Dobbe.
\newblock Concrete {Problems} in {AI} {Safety}, {Revisited}, December 2023.
\newblock URL \url{http://arxiv.org/abs/2401.10899}.
\newblock arXiv:2401.10899.

\bibitem[Raji and Yang(2020)]{raji_about_2020}
Inioluwa~Deborah Raji and Jingying Yang.
\newblock {ABOUT} {ML}: {Annotation} and {Benchmarking} on {Understanding} and
  {Transparency} of {Machine} {Learning} {Lifecycles}, January 2020.
\newblock URL \url{http://arxiv.org/abs/1912.06166}.
\newblock arXiv:1912.06166.

\bibitem[Raji et~al.(2020)Raji, Smart, White, Mitchell, Gebru, Hutchinson,
  Smith-Loud, Theron, and Barnes]{raji_closing_2020}
Inioluwa~Deborah Raji, Andrew Smart, Rebecca~N. White, Margaret Mitchell,
  Timnit Gebru, Ben Hutchinson, Jamila Smith-Loud, Daniel Theron, and Parker
  Barnes.
\newblock Closing the {AI} {Accountability} {Gap}: {Defining} an {End}-to-{End}
  {Framework} for {Internal} {Algorithmic} {Auditing}, January 2020.
\newblock URL \url{http://arxiv.org/abs/2001.00973}.
\newblock arXiv:2001.00973.

\bibitem[Raji et~al.(2021)Raji, Bender, Paullada, Denton, and
  Hanna]{raji_ai_2021}
Inioluwa~Deborah Raji, Emily~M. Bender, Amandalynne Paullada, Emily Denton, and
  Alex Hanna.
\newblock {AI} and the {Everything} in the {Whole} {Wide} {World} {Benchmark},
  November 2021.
\newblock URL \url{http://arxiv.org/abs/2111.15366}.
\newblock arXiv:2111.15366.

\bibitem[Raji et~al.(2022{\natexlab{a}})Raji, Kumar, Horowitz, and
  Selbst]{raji_fallacy_2022}
Inioluwa~Deborah Raji, I.~Elizabeth Kumar, Aaron Horowitz, and Andrew~D.
  Selbst.
\newblock The {Fallacy} of {AI} {Functionality}, July 2022{\natexlab{a}}.
\newblock URL \url{http://arxiv.org/abs/2206.09511}.
\newblock arXiv:2206.09511.

\bibitem[Raji et~al.(2022{\natexlab{b}})Raji, Xu, Honigsberg, and
  Ho]{raji_outsider_2022}
Inioluwa~Deborah Raji, Peggy Xu, Colleen Honigsberg, and Daniel~E. Ho.
\newblock Outsider {Oversight}: {Designing} a {Third} {Party} {Audit}
  {Ecosystem} for {AI} {Governance}, June 2022{\natexlab{b}}.
\newblock URL \url{http://arxiv.org/abs/2206.04737}.
\newblock arXiv:2206.04737.

\bibitem[Rajpurkar et~al.(2016)Rajpurkar, Zhang, Lopyrev, and
  Liang]{rajpurkar_squad:_2016}
Pranav Rajpurkar, Jian Zhang, Konstantin Lopyrev, and Percy Liang.
\newblock {SQuAD}: 100,000+ {Questions} for {Machine} {Comprehension} of
  {Text}, October 2016.
\newblock URL \url{http://arxiv.org/abs/1606.05250}.
\newblock arXiv:1606.05250.

\bibitem[Rauh et~al.(2024)Rauh, Marchal, Manzini, Hendricks, Comanescu,
  Akbulut, Stepleton, Mateos-Garcia, Bergman, Kay, Griffin, Bariach, Gabriel,
  Rieser, Isaac, and Weidinger]{rauh_gaps_2024}
Maribeth Rauh, Nahema Marchal, Arianna Manzini, Lisa~Anne Hendricks, Ramona
  Comanescu, Canfer Akbulut, Tom Stepleton, Juan Mateos-Garcia, Stevie Bergman,
  Jackie Kay, Conor Griffin, Ben Bariach, Iason Gabriel, Verena Rieser, William
  Isaac, and Laura Weidinger.
\newblock Gaps in the {Safety} {Evaluation} of {Generative} {AI}.
\newblock \emph{Proceedings of the AAAI/ACM Conference on AI, Ethics, and
  Society}, 7\penalty0 (1):\penalty0 1200--1217, October 2024.
\newblock ISSN 3065-8365.
\newblock \doi{10.1609/aies.v7i1.31717}.
\newblock URL \url{https://ojs.aaai.org/index.php/AIES/article/view/31717}.

\bibitem[Reid(2024)]{reid_generative_2024}
Liz Reid.
\newblock Generative {AI} in {Search}: {Let} {Google} do the searching for you,
  May 2024.
\newblock URL
  \url{https://blog.google/products/search/generative-ai-google-search-may-2024/}.

\bibitem[Ribeiro et~al.(2020)Ribeiro, Wu, Guestrin, and
  Singh]{ribeiro_beyond_2020}
Marco~Tulio Ribeiro, Tongshuang Wu, Carlos Guestrin, and Sameer Singh.
\newblock Beyond {Accuracy}: {Behavioral} {Testing} of {NLP} models with
  {CheckList}, May 2020.
\newblock URL \url{http://arxiv.org/abs/2005.04118}.
\newblock arXiv:2005.04118.

\bibitem[Rismani et~al.(2023)Rismani, Shelby, Smart, Jatho, Kroll, Moon, and
  Rostamzadeh]{rismani_plane_2023}
Shalaleh Rismani, Renee Shelby, Andrew Smart, Edgar Jatho, Joshua Kroll, AJung
  Moon, and Negar Rostamzadeh.
\newblock From {Plane} {Crashes} to {Algorithmic} {Harm}: {Applicability} of
  {Safety} {Engineering} {Frameworks} for {Responsible} {ML}.
\newblock In \emph{Proceedings of the 2023 {CHI} {Conference} on {Human}
  {Factors} in {Computing} {Systems}}, {CHI} '23, pages 1--18, New York, NY,
  USA, April 2023. Association for Computing Machinery.
\newblock ISBN 9781450394215.
\newblock \doi{10.1145/3544548.3581407}.
\newblock URL \url{https://dl.acm.org/doi/10.1145/3544548.3581407}.

\bibitem[Roose(2023)]{roose_conversation_2023}
Kevin Roose.
\newblock A {Conversation} {With} {Bing}’s {Chatbot} {Left} {Me} {Deeply}
  {Unsettled}.
\newblock \emph{The New York Times}, February 2023.
\newblock URL
  \url{https://www.nytimes.com/2023/02/16/technology/bing-chatbot-microsoft-chatgpt.html}.

\bibitem[Roose(2024{\natexlab{a}})]{roose_.i._2024}
Kevin Roose.
\newblock A.{I}. {Has} a {Measurement} {Problem}.
\newblock \emph{The New York Times}, April 2024{\natexlab{a}}.
\newblock URL
  \url{https://www.nytimes.com/2024/04/15/technology/ai-models-measurement.html}.

\bibitem[Roose(2024{\natexlab{b}})]{roose_can_2024}
Kevin Roose.
\newblock Can {A}.{I}. {Be} {Blamed} for a {Teen}’s {Suicide}?
\newblock \emph{The New York Times}, October 2024{\natexlab{b}}.
\newblock URL
  \url{https://www.nytimes.com/2024/10/23/technology/characterai-lawsuit-teen-suicide.html}.

\bibitem[Saxon et~al.(2024)Saxon, Holtzman, West, Wang, and
  Saphra]{saxon_benchmarks_2024}
Michael Saxon, Ari Holtzman, Peter West, William~Yang Wang, and Naomi Saphra.
\newblock Benchmarks as {Microscopes}: {A} {Call} for {Model} {Metrology}, July
  2024.
\newblock URL \url{http://arxiv.org/abs/2407.16711}.
\newblock arXiv:2407.16711.

\bibitem[Selbst et~al.(2019)Selbst, Boyd, Friedler, Venkatasubramanian, and
  Vertesi]{selbst_fairness_2019}
Andrew~D. Selbst, Danah Boyd, Sorelle~A. Friedler, Suresh Venkatasubramanian,
  and Janet Vertesi.
\newblock Fairness and {Abstraction} in {Sociotechnical} {Systems}.
\newblock In \emph{Proceedings of the {Conference} on {Fairness},
  {Accountability}, and {Transparency}}, {FAT}* '19, pages 59--68, New York,
  NY, USA, January 2019. Association for Computing Machinery.
\newblock ISBN 9781450361255.
\newblock \doi{10.1145/3287560.3287598}.
\newblock URL \url{https://dl.acm.org/doi/10.1145/3287560.3287598}.

\bibitem[Shiffrin and Mitchell(2023)]{shiffrin_probing_2023}
Richard Shiffrin and Melanie Mitchell.
\newblock Probing the psychology of {AI} models.
\newblock \emph{Proceedings of the National Academy of Sciences}, 120\penalty0
  (10):\penalty0 e2300963120, March 2023.
\newblock ISSN 0027-8424, 1091-6490.
\newblock \doi{10.1073/pnas.2300963120}.
\newblock URL \url{https://pnas.org/doi/10.1073/pnas.2300963120}.

\bibitem[Singer(2023{\natexlab{a}})]{singer_classrooms_2023}
Natasha Singer.
\newblock In {Classrooms}, {Teachers} {Put} {A}.{I}. {Tutoring} {Bots} to the
  {Test}.
\newblock \emph{The New York Times}, June 2023{\natexlab{a}}.
\newblock URL
  \url{https://www.nytimes.com/2023/06/26/technology/newark-schools-khan-tutoring-ai.html}.

\bibitem[Singer(2023{\natexlab{b}})]{singer_new_2023}
Natasha Singer.
\newblock New {A}.{I}. {Chatbot} {Tutors} {Could} {Upend} {Student} {Learning}.
\newblock \emph{The New York Times}, June 2023{\natexlab{b}}.
\newblock URL
  \url{https://www.nytimes.com/2023/06/08/business/khan-ai-gpt-tutoring-bot.html}.

\bibitem[Solaiman et~al.(2024)Solaiman, Talat, Agnew, Ahmad, Baker, Blodgett,
  Chen, III, Dodge, Duan, Evans, Friedrich, Ghosh, Gohar, Hooker, Jernite,
  Kalluri, Lusoli, Leidinger, Lin, Lin, Luccioni, Mickel, Mitchell, Newman,
  Ovalle, Png, Singh, Strait, Struppek, and
  Subramonian]{solaiman_evaluating_2024}
Irene Solaiman, Zeerak Talat, William Agnew, Lama Ahmad, Dylan Baker, Su~Lin
  Blodgett, Canyu Chen, Hal~Daumé III, Jesse Dodge, Isabella Duan, Ellie
  Evans, Felix Friedrich, Avijit Ghosh, Usman Gohar, Sara Hooker, Yacine
  Jernite, Ria Kalluri, Alberto Lusoli, Alina Leidinger, Michelle Lin, Xiuzhu
  Lin, Sasha Luccioni, Jennifer Mickel, Margaret Mitchell, Jessica Newman,
  Anaelia Ovalle, Marie-Therese Png, Shubham Singh, Andrew Strait, Lukas
  Struppek, and Arjun Subramonian.
\newblock Evaluating the {Social} {Impact} of {Generative} {AI} {Systems} in
  {Systems} and {Society}, June 2024.
\newblock URL \url{http://arxiv.org/abs/2306.05949}.
\newblock arXiv:2306.05949.

\bibitem[Stokes(1997)]{stokes_pasteurs_1997}
Donald~E. Stokes.
\newblock \emph{Pasteur’s {Quadrant}}.
\newblock Brookings Institution Press, Washington, D.C., 1997.

\bibitem[Timmermans and Berg(2003)]{timmermans_gold_2003}
Stefan Timmermans and Marc Berg.
\newblock \emph{The {Gold} {Standard}: {The} {Challenge} of {Evidence}-{Based}
  {Medicine} and {Standardization} in {Health} {Care}}.
\newblock Temple University Press, Philadelphia, PA, 2003.

\bibitem[{United Nations}(2011)]{united_nations_proposal_2011}
{United Nations}.
\newblock Proposal to develop amendments to global technical regulation {No}. 9
  concerning pedestrian safety., 2011.
\newblock URL
  \url{https://unece.org/fileadmin/DAM/trans/main/wp29/wp29wgs/wp29gen/wp29glob/ECE-TRANS-180-Add9-Amend1-App1e.pdf}.

\bibitem[{U.S. Food \& Drug Administration}(2017)]{health_evaluation_2017}
{U.S. Food \& Drug Administration}.
\newblock Evaluation and {Reporting} of {Age}-, {Race}-, and
  {Ethnicity}-{Specific} {Data} in {Medical} {Device} {Clinical} {Studies},
  September 2017.
\newblock URL
  \url{https://www.fda.gov/regulatory-information/search-fda-guidance-documents/evaluation-and-reporting-age-race-and-ethnicity-specific-data-medical-device-clinical-studies}.

\bibitem[Vedung(2017)]{vedung_public_2017}
Evert Vedung.
\newblock \emph{Public policy and program evaluation}.
\newblock Routledge, New York, 2017.

\bibitem[Verma(2023)]{verma_they_2023}
Pranshu Verma.
\newblock They fell in love with {AI} bots. {A} software update broke their
  hearts.
\newblock \emph{The Washington Post}, March 2023.
\newblock URL
  \url{https://www.washingtonpost.com/technology/2023/03/30/replika-ai-chatbot-update/}.

\bibitem[Wagner et~al.(2021)Wagner, Strohmaier, Olteanu, Kıcıman, Contractor,
  and Eliassi-Rad]{wagner_measuring_2021}
Claudia Wagner, Markus Strohmaier, Alexandra Olteanu, Emre Kıcıman, Noshir
  Contractor, and Tina Eliassi-Rad.
\newblock Measuring algorithmically infused societies.
\newblock \emph{Nature}, 595\penalty0 (7866):\penalty0 197--204, July 2021.
\newblock ISSN 1476-4687.
\newblock \doi{10.1038/s41586-021-03666-1}.
\newblock URL \url{https://www.nature.com/articles/s41586-021-03666-1}.

\bibitem[Wallach et~al.(2024)Wallach, Desai, Pangakis, Cooper, Wang, Barocas,
  Chouldechova, Atalla, Blodgett, Corvi, Dow, Garcia-Gathright, Olteanu, Reed,
  Sheng, Vann, Vaughan, Vogel, Washington, and Jacobs]{wallach_evaluating_2024}
Hanna Wallach, Meera Desai, Nicholas Pangakis, A.~Feder Cooper, Angelina Wang,
  Solon Barocas, Alexandra Chouldechova, Chad Atalla, Su~Lin Blodgett, Emily
  Corvi, P.~Alex Dow, Jean Garcia-Gathright, Alexandra Olteanu, Stefanie Reed,
  Emily Sheng, Dan Vann, Jennifer~Wortman Vaughan, Matthew Vogel, Hannah
  Washington, and Abigail~Z. Jacobs.
\newblock Evaluating {Generative} {AI} {Systems} is a {Social} {Science}
  {Measurement} {Challenge}, November 2024.
\newblock URL \url{http://arxiv.org/abs/2411.10939}.
\newblock arXiv:2411.10939.

\bibitem[Wang et~al.(2019)Wang, Singh, Michael, Hill, Levy, and
  Bowman]{wang_glue:_2019}
Alex Wang, Amanpreet Singh, Julian Michael, Felix Hill, Omer Levy, and
  Samuel~R. Bowman.
\newblock {GLUE}: {A} {Multi}-{Task} {Benchmark} and {Analysis} {Platform} for
  {Natural} {Language} {Understanding}, February 2019.
\newblock URL \url{http://arxiv.org/abs/1804.07461}.
\newblock arXiv:1804.07461.

\bibitem[Wang et~al.(2024)Wang, Hertzmann, and Russakovsky]{wang2024benchmark}
Angelina Wang, Aaron Hertzmann, and Olga Russakovsky.
\newblock Benchmark suites instead of leaderboards for evaluating ai fairness.
\newblock \emph{Patterns}, 5\penalty0 (11), 2024.

\bibitem[Weidinger et~al.(2023)Weidinger, Rauh, Marchal, Manzini, Hendricks,
  Mateos-Garcia, Bergman, Kay, Griffin, Bariach, Gabriel, Rieser, and
  Isaac]{weidinger_sociotechnical_2023}
Laura Weidinger, Maribeth Rauh, Nahema Marchal, Arianna Manzini, Lisa~Anne
  Hendricks, Juan Mateos-Garcia, Stevie Bergman, Jackie Kay, Conor Griffin, Ben
  Bariach, Iason Gabriel, Verena Rieser, and William Isaac.
\newblock Sociotechnical {Safety} {Evaluation} of {Generative} {AI} {Systems},
  October 2023.
\newblock URL \url{http://arxiv.org/abs/2310.11986}.
\newblock arXiv:2310.11986.

\bibitem[{White House}(2025)]{whitehouse_removing_2025}
{White House}.
\newblock Removing {Barriers} to {American} {Leadership} in {Artificial}
  {Intelligence}, January 2025.
\newblock URL
  \url{https://www.whitehouse.gov/presidential-actions/2025/01/removing-barriers-to-american-leadership-in-artificial-intelligence/}.

\bibitem[Wiggers(2024)]{wiggers_ai_2024}
Kyle Wiggers.
\newblock The {AI} industry is obsessed with {Chatbot} {Arena}, but it might
  not be the best benchmark, September 2024.
\newblock URL
  \url{https://techcrunch.com/2024/09/05/the-ai-industry-is-obsessed-with-chatbot-arena-but-it-might-not-be-the-best-benchmark/}.

\bibitem[Wimsatt(1994)]{wimsatt_ontology_1994}
William~C. Wimsatt.
\newblock The {Ontology} of {Complex} {Systems}: {Levels} of {Organization},
  {Perspectives}, and {Causal} {Thickets}.
\newblock \emph{Canadian Journal of Philosophy Supplementary Volume},
  20:\penalty0 207--74, 1994.

\bibitem[Zhang et~al.(2024)Zhang, Da, Lee, Robinson, Wu, Song, Zhao, Raja,
  Zhuang, Slack, Lyu, Hendryx, Kaplan, Lunati, and Yue]{zhang_careful_2024}
Hugh Zhang, Jeff Da, Dean Lee, Vaughn Robinson, Catherine Wu, Will Song,
  Tiffany Zhao, Pranav Raja, Charlotte Zhuang, Dylan Slack, Qin Lyu, Sean
  Hendryx, Russell Kaplan, Michele Lunati, and Summer Yue.
\newblock A {Careful} {Examination} of {Large} {Language} {Model} {Performance}
  on {Grade} {School} {Arithmetic}, November 2024.
\newblock URL \url{http://arxiv.org/abs/2405.00332}.
\newblock arXiv:2405.00332.

\bibitem[Zhou et~al.(2024)Zhou, Kim, Brahman, Jiang, Zhu, Lu, Xu, Lin, Choi,
  Mireshghallah, Bras, and Sap]{zhou_haicosystem:_2024}
Xuhui Zhou, Hyunwoo Kim, Faeze Brahman, Liwei Jiang, Hao Zhu, Ximing Lu, Frank
  Xu, Bill~Yuchen Lin, Yejin Choi, Niloofar Mireshghallah, Ronan~Le Bras, and
  Maarten Sap.
\newblock {HAICOSYSTEM}: {An} {Ecosystem} for {Sandboxing} {Safety} {Risks} in
  {Human}-{AI} {Interactions}, October 2024.
\newblock URL \url{http://arxiv.org/abs/2409.16427}.
\newblock arXiv:2409.16427.

\end{thebibliography}

\end{document}